\documentclass[lettersize,journal]{IEEEtran}
\usepackage{geometry}                		% See geometry.pdf to learn the layout options. There are lots.
\usepackage[parfill]{parskip}    		        % Activate to begin paragraphs with an empty line rather than an indent
\usepackage{graphicx}				% Use pdf, png, jpg, or eps§ with pdflatex; use eps in DVI mode
\usepackage{amssymb}				% !TEX TS-program = latex   pdflatexmk
\usepackage{amsmath}
\usepackage{xcolor}
\usepackage{textcomp}
\usepackage{algorithm} 
\usepackage{algpseudocode}
\usepackage{bbm}
\usepackage{tabularx}
\usepackage{latexsym}
\usepackage{subcaption}
\usepackage{gensymb}
\usepackage{fancyhdr}
\usepackage{eucal}
\usepackage{float}
\usepackage{parskip}
\usepackage{textcomp}

\title{Neuromorphic FPGA Design for Digital Signal Processing}  
\usepackage{geometry}                		% See geometry.pdf to learn the layout options. There are lots.
\usepackage{tabularx}
\usepackage{mathrsfs}
\usepackage{calrsfs}
\usepackage{stmaryrd}
\usepackage{wasysym} 
\usepackage{xurl}
\newcommand{\code}[1]{\texttt{#1}}

\usepackage{microtype}
\usepackage{yfonts}
\usepackage{listings} 
\usepackage{multirow}
\usepackage{xcolor}
\usepackage{pdfpages}
\usepackage{graphicx} 
\usepackage[utf8]{inputenc}
\usepackage [english]{babel}
\usepackage [autostyle, english = american]{csquotes}
\usepackage[style=numeric,natbib=true]{biblatex} 
\graphicspath{{./Images/}} 
\usepackage{mathtools}
\usepackage{amsmath}
\usepackage{fancyhdr} 
\usepackage{titlesec}
\usepackage{upgreek} 
\usepackage{amsmath} 
\usepackage{amssymb}

\PassOptionsToPackage{hyphens, spaces, obeyspaces}{url} 
\usepackage[final=true,backref=true,colorlinks=true,linkcolor=blue, urlcolor=blue, citecolor=blue, breaklinks=true]{hyperref}
\hypersetup{
    pagebackref=true,
    colorlinks=true,
    citecolor=blue,
    linkcolor=blue,
    filecolor=magenta,      
    urlcolor=blue,
}
\hypersetup{breaklinks=true}
\usepackage{breakurl}
\def\xHyphenate#1#2\wholeString {\if#1$%
    \else\transform{#1}%
    \takeTheRest#2\ofTheString\fi}

\def\takeTheRest#1\ofTheString\fi
{\fi \xHyphenate#1\wholeString}

\def\transform#1{\url{#1}\hskip 0pt plus 1pt}

\definecolor{vgreen}{RGB}{104,180,104}
\definecolor{vblue}{RGB}{49,49,255}
\definecolor{orange}{RGB}{255,143,102}

\definecolor{dkgreen}{rgb}{0,0.6,0}
\definecolor{gray}{rgb}{0.5,0.5,0.5}
\definecolor{mauve}{rgb}{0.58,0,0.82}
\definecolor{cred}{rgb}{220,20,60}
\definecolor{lightsilver}{rgb}{0.96,0.96,0.98} 
\lstdefinelanguage{Verilog}{
            keywords=[1]{module, input, output, initial, begin, end, always, posedge, for, or, negedge, if, else, endmodule}, % First set of keywords
            keywordstyle=[1]\color{mauve},
            keywords=[2]{wire, parameter, integer, reg, integer, real, \$display, \$finish, signed}, % Second set of keywords
            keywordstyle=[2]\color{red}
}
\author{\IEEEauthorblockN{Justin London} \\
\IEEEauthorblockA{\textit{Dept. of Electrical Engineering and Computer Science} \\
\textit{University of North Dakota}\\
Grand Forks, North Dakota \\
justin.london@und.edu \\
}}

\begin{document}

\maketitle
\newpage 
\begin{abstract}
    In this paper, the foundations of neuromorphic computing, spiking neural networks (SNNs) and memristors, are analyzed and discussed. Neuromorphic computing is then applied to FPGA design for digital signal processing (DSP).  Finite impulse response (FIR) and infinite impulse response (IIR) filters are implemented with and without neuromorphic computing in Vivado using Verilog HDL.  The results suggest that neuromorphic computing can provide low-latency and synaptic plasticity thereby enabling continuous on-chip learning. Due to their parallel and event-driven nature, neuromorphic computing can reduce power consumption by eliminating von Neumann bottlenecks and improve efficiency, but at the cost of reduced numeric precision.  
\end{abstract}
\section{Introduction}

\ \ \ Neuromorphic computing enables continuous on-chip learning and adaptation to new information without requiring extensive retraining using memristors.  Memristors or resistive switches are passive, nonlinear circuit elements with variable nonvolatile resistance states, similar to biological synapses. They are compatible with CMOS technology.  Memristor-based Spiking Neural Networks (SNNs) with temporal spike encoding enable ultra-low-energy computation, making them ideal for battery-powered intelligent devices \cite{Abreu:2024, Sune:2020}.  SNNs are building blocks for neuromorphic computing. In SNN circuit implementations, memristors enable the co-location of computation and memory, thus alleviating the memory bottleneck problem, a major impediment to data-intensive artificial neural network (ANN) workloads. They can be defined as the underlying computations performed by neuromorphic physical systems which carry out robust and efficient neural computation using hardware implementations that operate in physical time. Typically, they are event- or data-driven, and typically they employ low-power, massively parallel hybrid analog, digital or mixed VLSI circuits, and they operate using similar physics of computation used by the nervous system \cite{Mehonic:2024}.  
 
In particular, neuromorphic computing focuses on hardware systems that mimic the structure and function of the biological brain, typically involving neural networks and spiking neurons.  Neuromorphic FPGAs use CMOS devices and circuits to emulate neurons and synapses, and process data as spikes instead of continuous inputs \cite{Lu:2020}.   Memristors and SNNs have been used in deep learning neuromorphic FPGA designs \cite{Hong:2020}.  The main components of a memristive SNN include a include a temporal spike encoder, dual switches, a memristive crossbar (MC), Leaky Integrate-and-Fire (LIF) neurons, and various proposed control modules such as the lateral inhibition circuit (LIC), synapse control circuit (SCC), dual switch control circuit (DCC), and update control circuit (UCC) \cite{Prajapati:2025}   

Spikes that are distributed in space can be parallel processed, and those that are triggered across different times can be serialized.   In the absence of spiking events, less memory access are required, though computing spike-based workloads still benefit from parallel processing.  The sparsity of spike-based activations results in large sub-blocks of tensors remaining inactive in GPUs, which are not effectively optimized for sparsity, nor can they be reprogrammed to be so \cite{Karamimanesh:2025}.  von Neumann architecture prevalent in CPU and GPU processing enables data transfers between the processing unit and memory unit can lead to high power dissipation.  FPGAs provide highly parallel processing, high flexibility, and are reprogrammable; and therefore are suitable for spiking neural networks used in neuromorphic computing which can be applied to digital signal processing and filtering.

Finite impulse response (FIR) filters are non-recursive (feed-forward only) and offer linear phase characteristics, a major advantage in many DSP applications. Infinite impulse response (IIR) filters are are recursive (using feedback paths) and can achieve sharper frequency responses with fewer resources than equivalent FIR filters \cite{Woods:2017}.  FPGAs are well-suited for DSP applications due to their inherent parallelism and dedicated DSP slices (e.g., Xilinx DSP48E elements) that efficiently perform multiply-accumulate (MAC) operations. Verilog can be used to describe the digital logic for these filters.  Since, memristors can perform both data storage and computation within the same physical location (in-memory computing), overcoming the \enquote{von Neumann bottleneck} that limits the speed and power efficiency of traditional computer architectures, they are particularly useful for accelerating the multiply-and-accumulate (MAC) operations central to many DSP and AI tasks.

This study will implement digital signal processing FIR and IIR low-pass filters using Verilog HDL for FPGA design/synthesis and LTSpice for circuit simulation.  MUX-based multipliers, adders, or distributed arithmetic techniques to optimize for area, power, and speed will be used in their design.  To improve efficiency and speed, multiplexer-based truncated multipliers and approximate adders can also be used in their design. \cite{Bali:2025}  The FIR and IIR will then be redesigned to make them neuromorphic.  This can involve replacing the output stage with a SNN such as a leaky integrate-and-fire neuron model.  The filter's output can serve as the synaptic input current to this neuron.  Instead of direct digital convolution, the filter can be implemented as a network of digital spiking neurons, where data is represented by spike times or rates. The data-in and data-out would then represent spike events or encoded signals.  Neurons only process information when they receive a spike.  This can dramatically reduces power consumption compared to traditional DSP filters that constantly perform computations using sequential logic and shift registers to store previous inputs and outputs needed. 

Standard FIR and IIR filters can be incorporated as neuromorphic components in an FPGA design.  We seek to first design standard FIR and IIR filters for an FPGA using Verilog. A neuromorphic-inspired DSP filter architecture using SNNs, that mimics the structure of neural networks to perform filtering tasks, which may be more efficient in processing than traditional FIR/IIR designs, will then be implemented.  The objective of this work is to show how neuromorphic processing can improve the efficiency and processing of DSP filters in FPGAs.  Traditional DSP uses standard multipliers. Neuromorphic memristor-based designs might use highly scalable parallel spike-based digital multipliers or other novel arithmetic units to reduce power and area consumption.  

In contrast to standard DSP filters, neuromorphic filters can use adaptive tap coefficients often incorporate on-chip learning, meaning the COEFF values change dynamically using algorithms like LMS (Least Mean Squares), requiring the coefficients to be stored in re-writable RAM instead of ROM or parameters.  SNNs can achieve high energy efficiency with lower storage requirements, as they use sparse, binary (0-1) signals instead of real-valued activations and feature maps. Since some SNN-based filers can be designed with self-adaptive time windows, they can in principle be more robust to handling noise compared to traditional filters.

Unlike traditional processors that require constant clock cycles and data movement, neuromorphic chips compute only when a signal is received, process data closer to where it is stored (in-memory computing), and use analog computation to perform mathematical operations more efficiently.  Thus, neuromorphic computing can improve DSP performance by leveraging its event-driven, massively parallel, and memory-centric architecture, which leads to lower power consumption and higher efficiency for certain tasks like sensor processing and inference.

The FIR And IIR algorithms use signed integers to represent Q-format fixed-point numbers with integer and fractional bits. Default parameters ($DATA\_W = 24$, FRAC $\approx 16$) give a good mid-ground. One must choose FRAC consistently across signals (coefficients, data, thresholds).  Memristors can be used in DSP and neural computing for non-volatile memory
as they retain their resistance states (memopry) without power, enabling energy-efficient in-memory computing and mimicking biological synaptic plasticity.

This paper is organized as follows.  In section \ref{methodology}, the methodology used is detailed.  The methodology covers a background discussion of leaky integrate-and-fire models, FIR filters, and IIR Biquad  filters, respectively. In \ref{memristors}, the details of memristors, the foundational elements for neuromorphic computing, are discussed.  Time registers and operational logic as well as time-domain modules used in DSP are provided.   In section section \ref{FPGA}, FPGA design for DSP is covered.  In section \ref{results}, the results of the simulation are provided and analyzed.  Finally, section \ref{conclusion} concludes. 

\section{Methodology}
\label{methodology}

    In this study, Xilinx Vivado/Verilog HDL and LTSpice are used to implement and simulate waveforms and schematics for FIR filters, IIR filters, and memristors/neuromorphic filter models for digital signal processing.  We use a leak integrate-and-fire neuron (LIF) model.    A leaky integrate-and-fire neuron models a neuron's behavior by integrating input currents and, when the membrane potential crosses a threshold, "firing" a spike and resetting. The "leaky" part comes from an exponential decay of the membrane potential toward a resting potential, similar to an RC circuit.  Incoming signals from other neurons are modeled as current pulses that add to the neuron's total membrane potential where the membrane has leaky potential  At the same time, the membrane potential naturally "leaks" back towards a resting potential.   The fundamental equation for the LIF that describes how a neuron's membrane potential changes over time is:
    \begin{equation}
        \tau \frac{dV}{dt} = -(V(t)-V_{rest}) + RI_{in}(t)
    \end{equation}  
    where $V(t)$ is the membrane potential, $\tau$ is the membrane time constant, $V_{rest}$ is the resting potential, $R$ is the membrane resistance, and $I_{in}(t)$ is the input current.  Let $C = \frac{\tau}{R}$   be the equivalent capacitance of the membrane. so that the LIF neruon can be written as:
    \begin{equation}
        I_{in}(t) = \frac{V(t)-V_{rest}}{R} + C\frac{dV}{dt}
    \end{equation}
    In FPGAs, this continuous function can be discretized using Euler integration resulting in hardware-friendly difference equations.
    There are various spike neural network models and variations such as the adaptive exponential integrate and fire (AdEx) model, Hodgkin-Huxley, and Izhikevich model \cite{Karamimanesh:2025}  Figure \ref{fig:nms} compares the neuron models:
      \begin{figure}[H]
	   \centering
	   %\begin{center} 
	   \includegraphics[width=1\linewidth]{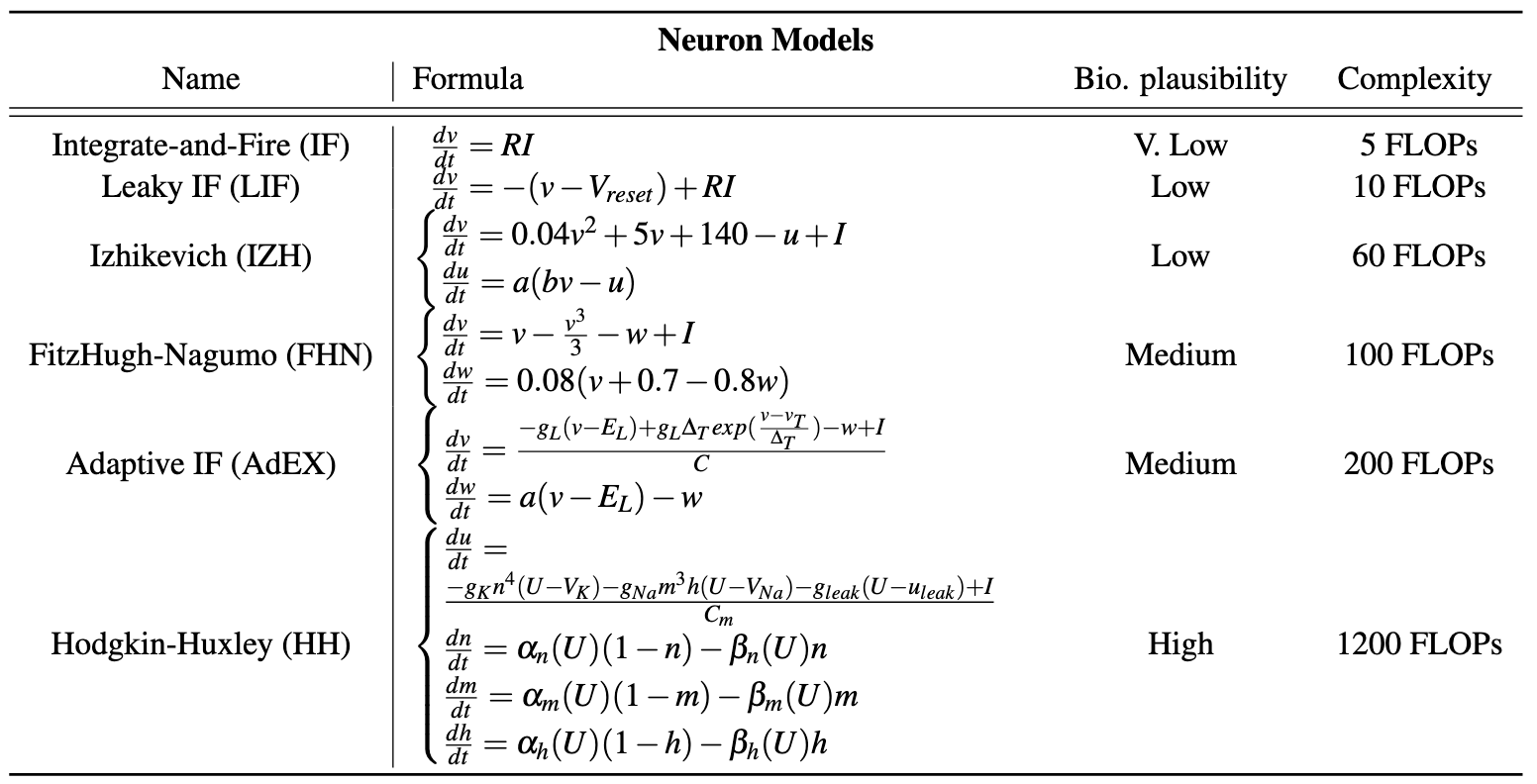}
	   %\small \caption{} 	
	   \caption{Comparison of Neuron Models. \cite{Szczerek:2025}}  
	   %\end{center}  
      \label{fig:nms} 
    \end{figure}

\begin{figure}[H]
	   \centering
	   %\begin{center} 
	   \includegraphics[width=1\linewidth]{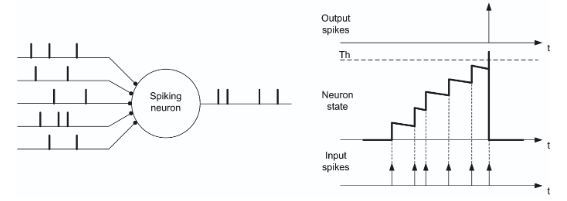}
	   %\small \caption{} 	
	   \caption{(left) Leaky integrate-and-fire spiking neuron; (right) temporal evolution of the neuron state while it receives input spikes. \cite{Camunas:2019}}  
	   %\end{center}  
      \label{fig:lif} 
    \end{figure}

\begin{figure}[H]
	   \centering
	   %\begin{center} 
	   \includegraphics[width=1\linewidth]{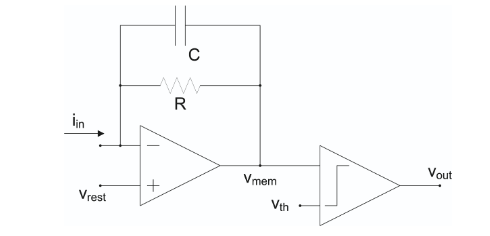}
	   %\small \caption{} 	
	   \caption{RC LIF neuron implementation. \cite{Camunas:2019}} 
	   %\end{center}  
      \label{fig:lif2} 
    \end{figure}

    Figure \ref{fig:lif} (left) illustrates the behavior of a LIF spiking neuron.  The spiking neuron receives spikes from several inputs, processes them, and generates output spikes from its output node. The temporal evolution of the neuron state while it receives input spikes is shown on the right.   When the threshold is reached, it generates and output spike.
    Figure \ref{fig:lif2} illustrates an RC leaky-and-integrate neuron hardware implementation.
    
    The neuron's decay is governed by a membrane time constant (\(\tau \)), which determines how quickly the potential drops.  If the integrated potential reaches a specific threshold voltage, the neuron "fires" an action potential (a spike).Resetting: Immediately after firing, the neuron's membrane potential is reset to a lower value (often the resting potential).  
    A brief, absolute refractory period follows the spike during which the neuron cannot fire again, though this is often handled simply by the reset mechanism.  A higher input current leads to a faster rise to the threshold, resulting in a higher firing rate.

    Spike timing-depending plasticity (STDP) is a common learning rule implemented on-chip.  The synaptic weight updates depend on the relative timing of pre- and post-synaptic spikes. The mathematical expression typically involves adding or subtracting a value from the synaptic weight based on the time difference, often using simple mask-and-accumulate operations instead of complex multiply-and-accumulate (MAC) as in traditional AI.   Hardware arithmetic implementations uses fixed-point arithmetic, reconfigurable multipliers, and spike encoding.

    Most FPGA-based neuromorphic systems use fixed-point number representation (e.g., 16-bit or 18-bit) for weights and membrane potentials because it uses fewer hardware resources (e.g., look-up tables (LUTs) and flip-flops (FFs)) and less power compared to floating-point.  FPGAs have embedded hardware multipliers (DSPs), but their use is optimized to handle the varied precision requirements of neuromorphic workloads, sometimes employing approximate computing techniques to balance accuracy and resource usage.  Information is primarily conveyed through binary spikes (event-driven), not continuous values, which simplifies the math to \enquote{mask-and-accumulate} or addition operations when a spike arrives.

\subsection{FIR Filters}
\label{fir}
    An FIR filter of length $M$ with input $x(n)$ and output $y(n)$:
\begin{equation}
    y(n) = \sum_{k=0}^{M-1}b_{k}x(n-k) 
\end{equation}
Alternatively, the output sequence can be expressed as a sequence as the convolution of the unit sample response $h(n)$ of the system with the input signal yielding
\begin{equation}
        y(n) = \sum_{k=0}^{M-1}h(k)x(n-k)   
\end{equation}
where $b_{k}=h(k)$.  The filter can be characterized by its system function
\begin{equation}
    H(z) = \sum_{k=0}^{M-1}h(k)z^{-k}
\end{equation}
where $z^{-k}$ is the lag of degree $k$

The frequency response is \(H(e^{j\omega })=\sum _{k=0}^{N-1}h[k]e^{-j\omega k}\) where \(h[k]\) are the filter coefficients.\(\omega \) is the frequency in radians per sample (ranging from \(0\) to \(\pi \) radians, where \(\pi \) is the Nyquist frequency.)  To compute the gain of an FIR filter, we use the formula for the frequency response, \(H(\omega )=\sum _{k=0}^{N-1}h[k]e^{-j\omega k}\), where \(h[k]\) are the filter coefficients, \(N\) is the filter length, and \(\omega \) is the frequency. The magnitude of this complex value, \(|H(\omega )|\), is the linear gain at that specific frequency. For a specific case like the DC gain (at 0 Hz), the gain is simply the sum of all coefficients: \(G_{DC}=\sum _{k=0}^{N-1}h[k]\).  To express the gain in decibels, we use the formula: \(Gain_{dB}=20\log _{10}(|H(\omega )|)\).  

FIR Digital Signal Processors (DSPs) primarily use dedicated multiplier-accumulate (MAC) units, along with adders, subtractors, registers (for pipelining and delay lines), and multiplexers. These are often combined into specialized, highly efficient "DSP blocks" within FPGAs or as hardwired components in standalone DSP chip.  Multipliers are often 48-bit or 64-bit registers that perform a series of multiply-add operations (Multiply-Accumulate, or MAC units) which is central to FIR filter operation.  Pipelining registers/shift registers are essential for the tap-delay lines in FIR filters. The input data is passed through a series of registers to create the necessary delays for each tap, allowing for high performance through pipelining.

\begin{figure}[H]
	\centering
	   %\begin{center} 
	\includegraphics[width=0.7\linewidth]{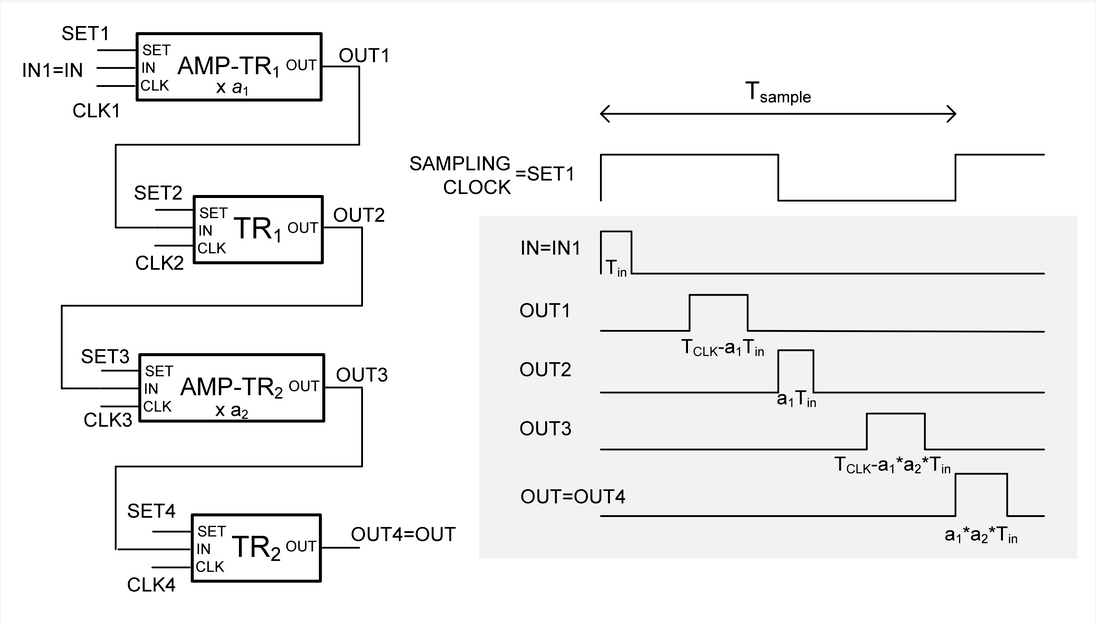}
	   %\small \caption{} 	
	\caption{(a) Time-mode $z^{-1}$ multiplier circuit, (b) pulses timing diagram. \cite{Felouris:2022}}  
	   %\end{center}  
    \label{fig:time1} 
\end{figure}

\begin{figure}[H]
	   \centering
	   %\begin{center} 
	   \includegraphics[width=1\linewidth]{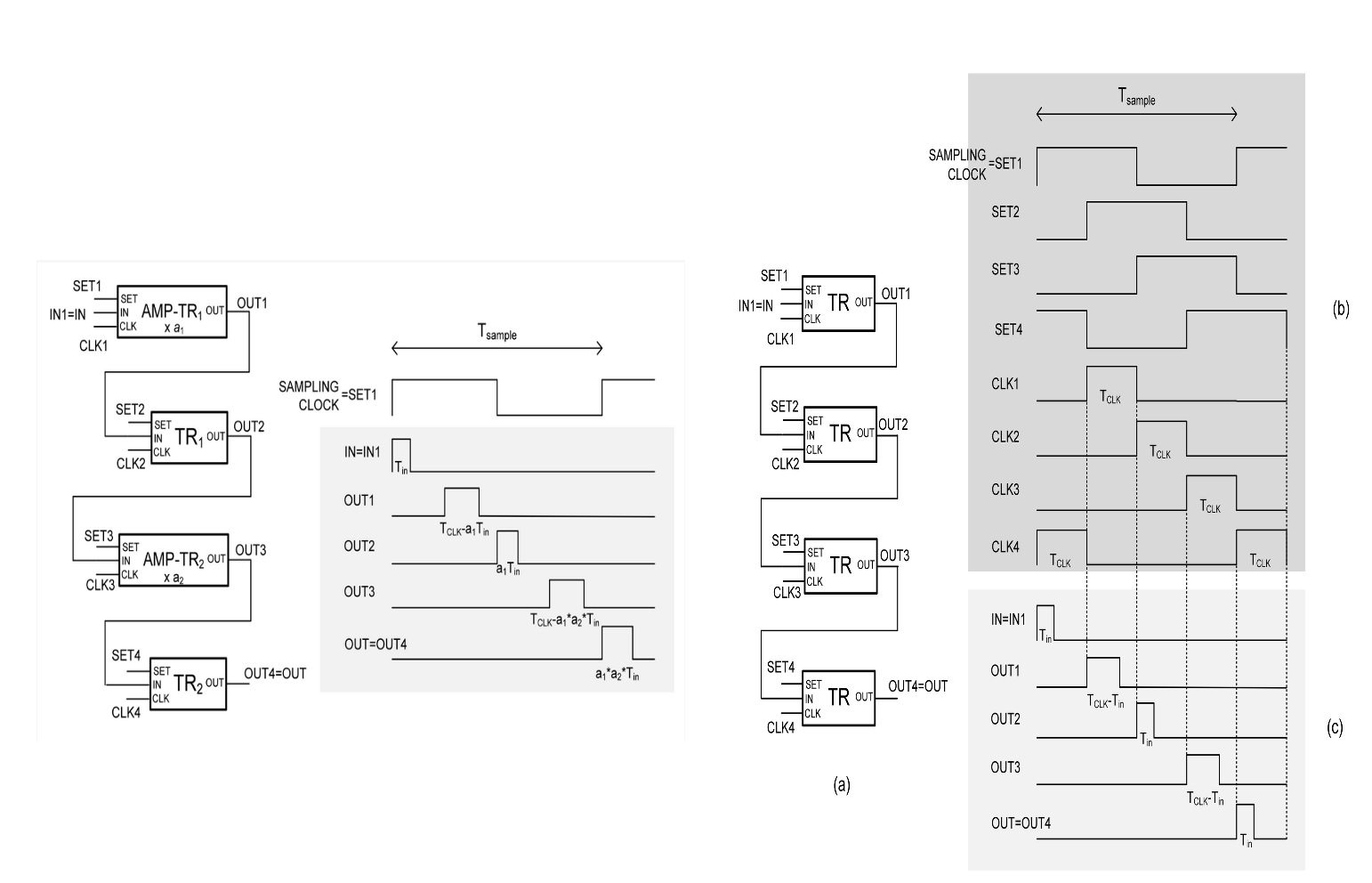}
	   %\small \caption{} 	
	   \caption{(left) (a) Time-mode $z^{-1}$ multiplier circuit; (b) pulse timing diagram; (right) (a) Time-domain $z^{-1}$ circuit, (b) clocks timing diagram , and (c) pulses timing diagram \cite{Felouris:2022}}.  
	   %\end{center}  
      \label{fig:op} 
\end{figure}

In DSP, a time-mode $z^{-1}$ operator multiplier, where $T_{in}(k)z^{-1} = T_{out}(k) = T_{in}(k-1)$, is required to produce an output pulse with pulse $T_{out}$ width equal to $a \cdot T_{in}$, where $T_{in}$ is the input pulse and $a$ is the multiplication coefficient, and the output pulses will be synchronized with the sampling signal and delayed by one clock cycle.  Figure \ref{fig:time1} shows the time-mode $z^{-1}$ multiplier circuit and timing pulses.  A time-mode $z^{-1}$ multiplier is realized using the combination of two AMP-TR and two TR circuit series as shown in Figure \ref{fig:op}. 

\subsection{IIR Biquad}
\label{iir}
\ \ \ The general formula for an infinite impulse response (IIR) filter is the difference equation:
\begin{equation}
    y[n] = \sum_{k=0}^{M} b_{k}x[n-ki] - \sum_{k=1}^{N} a_{k}y[n-k]
\end{equation}
representing the current output y[n] as a weighted sum of current/past inputs and past outputs using feedforward (b) and feedback (a) coefficients, or its transfer function:
\begin{equation}
    H(z) = \frac{B(x)}{A(z)} = \frac{\sum_{k=0}^{M} b_{k}z^{k} }{1 +\sum_{k=1}^{N}a_{k}z^{-k}}
\end{equation}

The term \enquote{biquad} is short for bi-quadratic and is a common name for a two-pole, two-zero digital filter.  The transfer function of the biquad can be defined as:
\begin{equation}
    H(z) = g\frac{1 + \beta_{1}z^{-1} + \beta_{2}z^{-2}}{1+a_{1}z^{-1}+a_{2}z^{-2}}
\end{equation}

\subsection{Butterworth Filter}
\label{butterworth}
\begin{figure}[H]
	   \centering
	   %\begin{center} 
	   \includegraphics[width=0.7\linewidth]{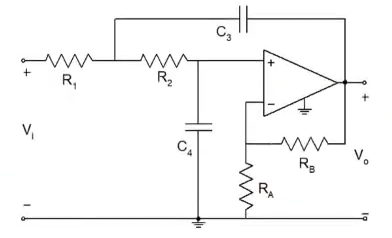}
	   %\small \caption{} 	
	   \caption{2nd order filter}  
	   %\end{center}  
      \label{fig:filter1} 
\end{figure}

\begin{figure}[H]
	   \centering
	   %\begin{center} 
	   \includegraphics[width=1\linewidth]{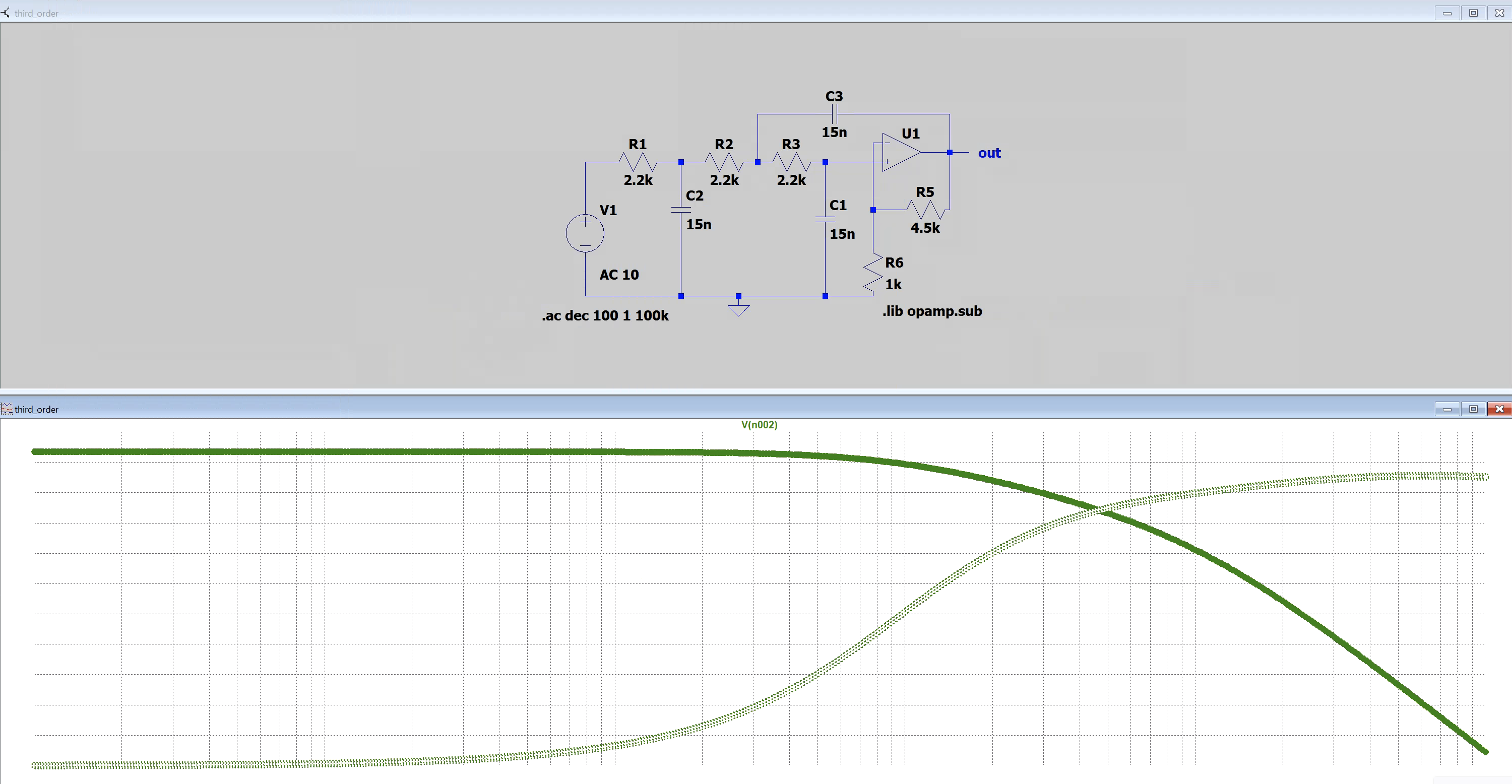}
	   %\small \caption{} 	
	   \caption{BW Circuit in LTSpice}  
	   %\end{center}  
      \label{fig:three1} 
\end{figure}

\begin{figure}[H]
	   \centering
	   %\begin{center} 
	   \includegraphics[width=1\linewidth]{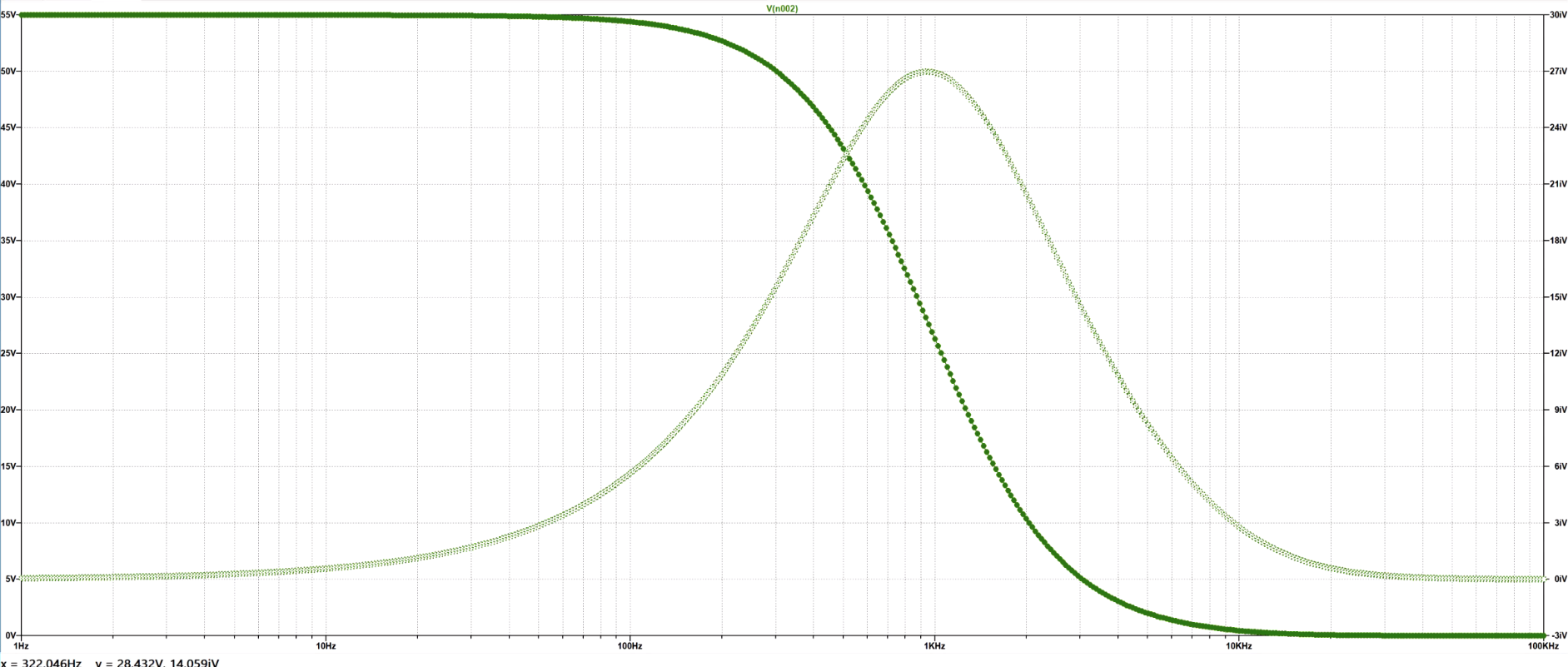}
	   %\small \caption{} 	
	   \caption{LTSpice Simulation of BW Filter}  
	   %\end{center}  
      \label{fig:three2} 
\end{figure}

    Consider a 2nd order Butterworth (Sallen-Key) low pass filter shown in Figure \ref{fig:filter1}.
    Assume $\alpha = 7$ and $\beta = 6$.  The DC gain $K = \frac{4 + \alpha}{2} = \frac{4 + 7}{2} = 5.5$.  The cut-off frequency = $f_{c} = \frac{4 + \beta}{2} = \frac{4 + 6}{2} = 5 K Hz$.  Let $f_{c} = \frac{1}{2 \pi \sqrt{R_{1}R_{2}C_{1}C_{2}}}$ be the cut-off frequency and assume $R = R_{1} = R_{2}$ and $C = C_{1}=C_{2} = 15 nF$ yielding $f_{c} = \frac{1}{2\pi RC}$ so that $R = \frac{1}{2 \pi f_{c}C} = \frac{1}{2 \pi (5 k Hz)(15 nF)} = 2.2 k \Omega$  The transfer function is
    \begin{multline}
        \frac{V_{o}(s)}{V_{i}(s)} = \\ 
        \frac{K\frac{1}{R_{1}R_{2}C_{3}C_{4}}}{s^2 + s \big (\frac{1}{R_{1}C_{3}} + \frac{1}{R_{2}C_{3}} + \frac{1}{R_{2}C_{4}} - \frac{K}{R_{2}C_{4}}  \big )  + \frac{1}{R_{1}{R_{2}C_{3}C_{4}}}} \nonumber
    \end{multline}
    or
    \begin{multline}
               \frac{K\frac{1}{R^{2}C^{2}}}{s^2 + s \big (\frac{1}{RC + \frac{1}{RC} + \frac{1}{RC} - \frac{K}{RC}} \big )  + \frac{1}{R^{2}C^{2}}}\nonumber
    \end{multline}
    Plugging in the values yields
    \begin{equation}
        \frac{V_{o}(s)}{V_{i}(s)} = \frac{5.05 \times 10^{9}}{s^2 - 75757.7s + 9.2 \times 10^{8}}
    \end{equation}
    where $K = 1 + \frac{R_{A}}{R_{B}}$.  If $R_{A} = 1 \ k \Omega$, then $R_{B} = 4.5 k\Omega$.  We show the BW low pass filter circuit and simulation in LTSpice as shown in Figure \ref{fig:three1} and \ref{fig:three2}, respectively.

\subsection{High Passfilter}

\begin{figure}[H]
	   \centering
	   %\begin{center} 
	   \includegraphics[width=1\linewidth]{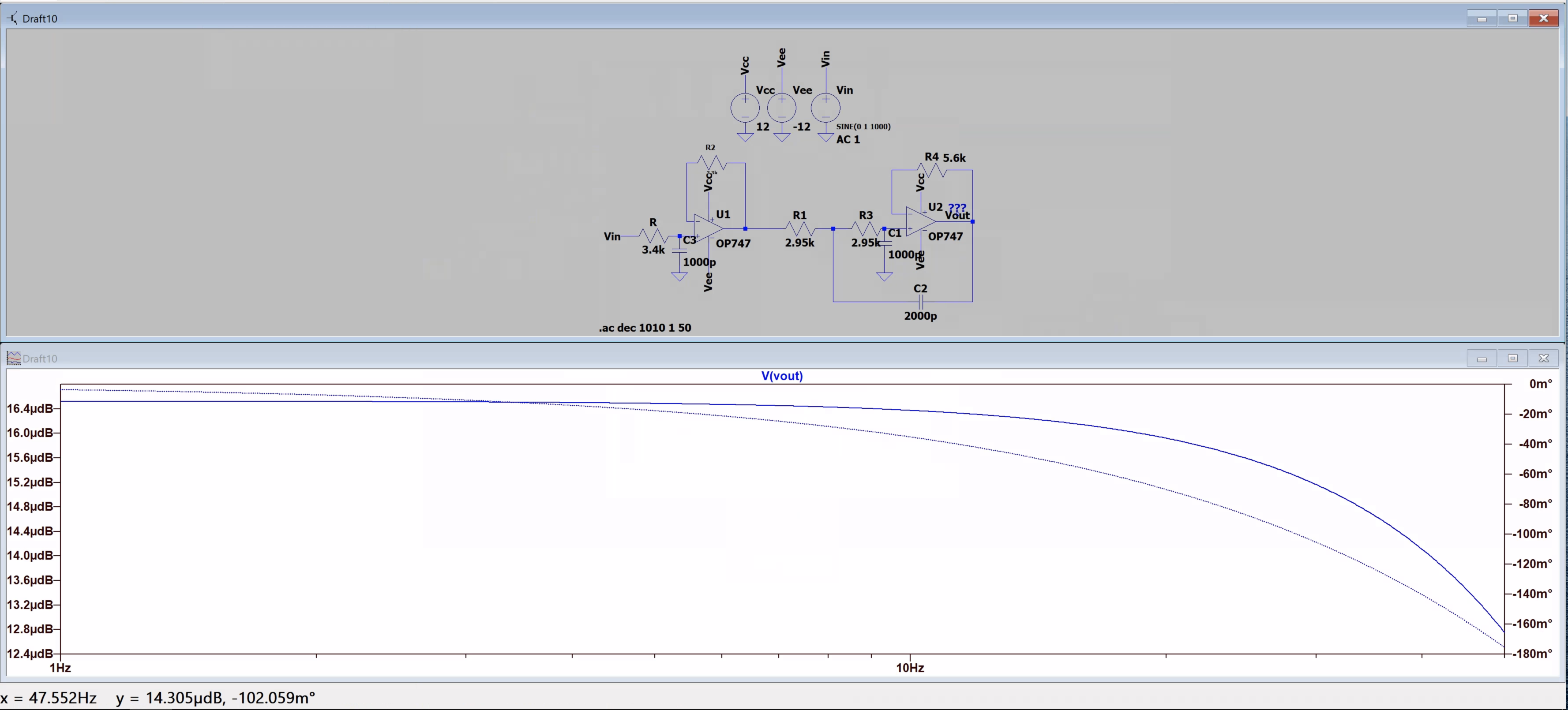}
	   %\small \caption{} 	
	   \caption{Highpass Filter}  
	   %\end{center}  
      \label{fig:high} 
\end{figure}

Figure \ref{fig:high} illustrates an LTSpice simulation for a third-order active high pass filter which uses two differential op amplifiers.

Figure \ref{fig:band} illustrates an LTSpice simulation for a passive bandpass filter:
\begin{figure}[H]
	   \centering
	   %\begin{center} 
	   \includegraphics[width=1\linewidth]{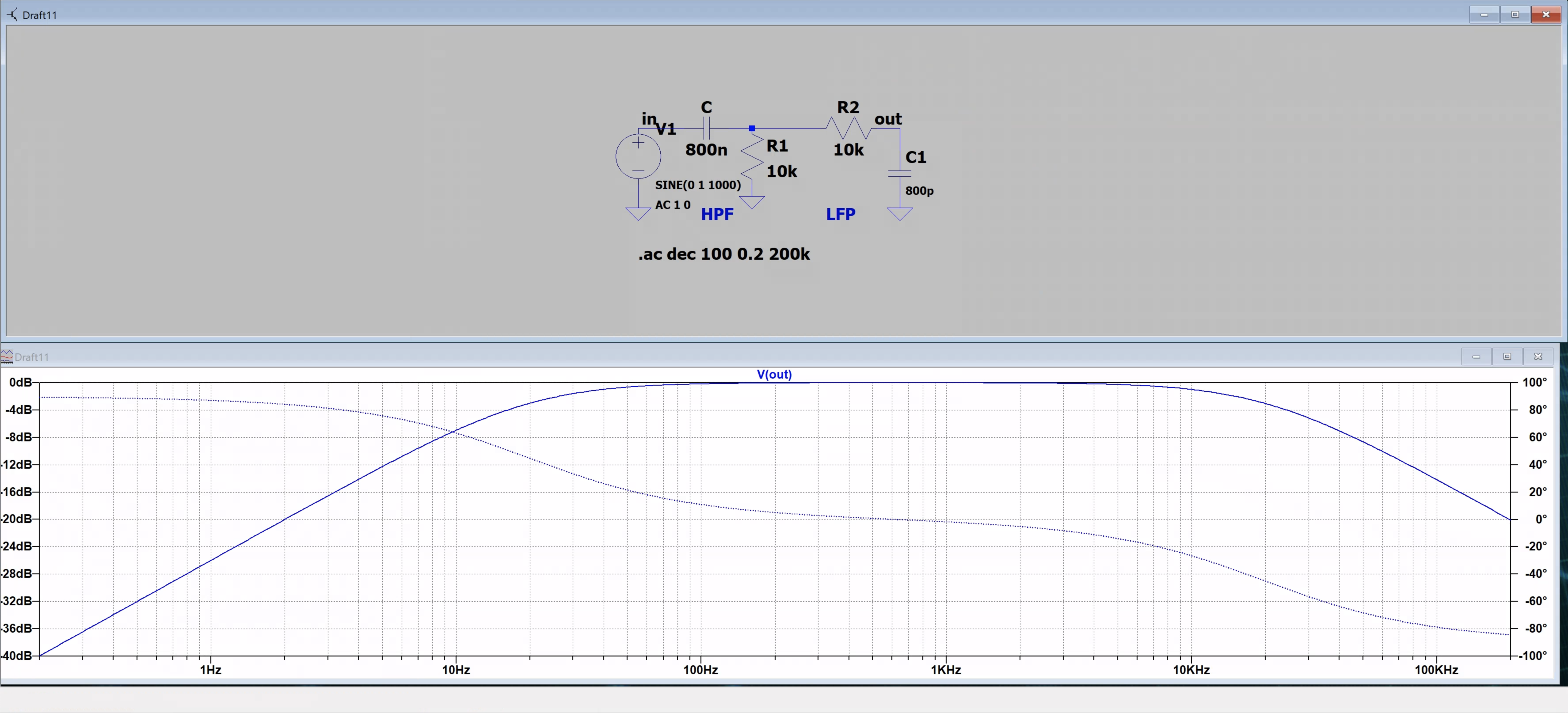}
	   %\small \caption{} 	
	   \caption{Bandpass Filter}  
	   %\end{center}  
      \label{fig:band} 
\end{figure}

\section{Memristors}
\label{memristors}

\begin{figure}[h]
	   \centering
	   %\begin{center} 
	   \includegraphics[width=1\linewidth]{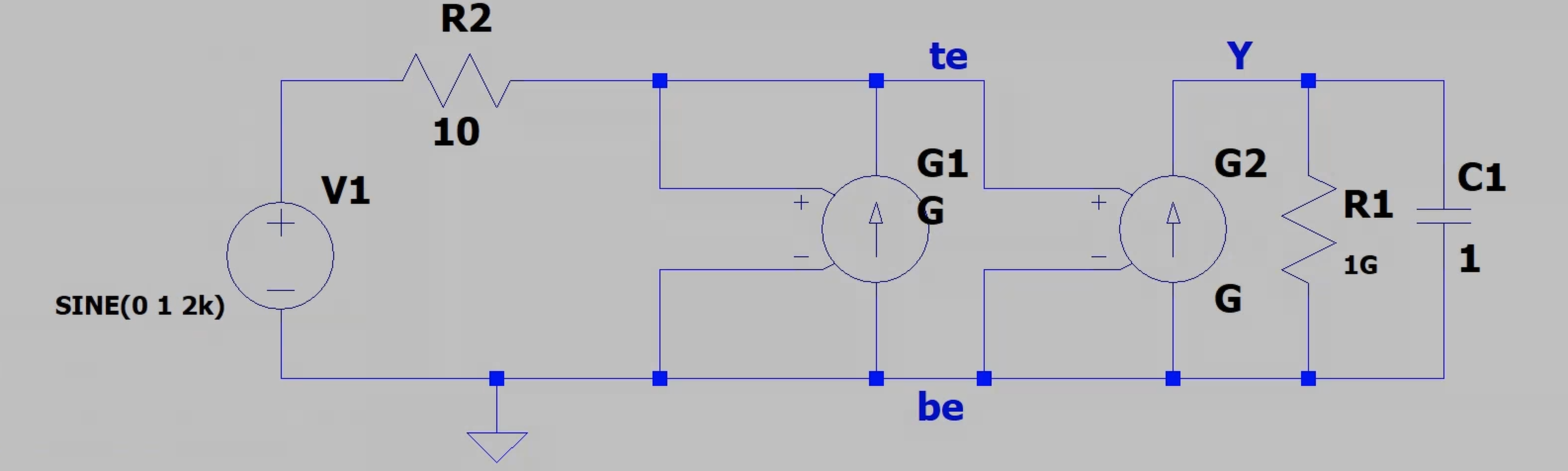}
	   %\small \caption{} 	
	   \caption{Memristor Circuit}  
	   %\end{center}  
      \label{fig:mem} 
\end{figure}

\begin{figure}[H]
	   \centering
	   %\begin{center} 
	   \includegraphics[width=1\linewidth]{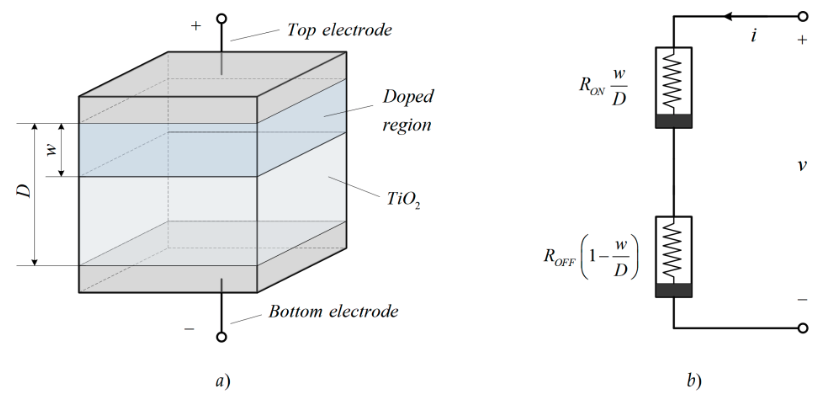}
	   %\small \caption{} 	
	   \caption{(a) Memristor nanostructure and (b) equivalent electric circuit. \cite{Mladenov:2021}}  
	   %\end{center}  
      \label{fig:structure} 
\end{figure}

A memristor-based neuron, two-terminal/electrode switches with variable conductance, has input signals $x_{1},\dots,x_{n}$ that are directed to synapses based on memristors, with corresponding weights $w_{1},\cdots,w_{N}$ \cite{Mladenov:2024}.   The change in the respective synaptic weight is realized by alteration of the corresponding memristance with externally applied voltage or current pulses.   They are have low power consumption, high switching speed, and good compatibility with complementary metal oxide semiconductor (CMOS) technology \cite{Mladenov:2021}.  Figure \ref{fig:mem} illustrates a schematic of a memristor.   The top electrode is denoted \code{te} and the bottom electrode is \code{be}.  A sinusoidal voltage source supplies the memristor equivalent circuit.  The memristor current is modeled by the voltage-controlled currrent source $G_{1}$.  The current of this dependent source is the memristor current.   The additional electrode for measuring the memresistor state variable $x$ is denoted $Y$ .  The time derivative of the state variable is proportional to the memresistor current and is experessed by the dependent, voltage-controlled current source $G_{2} \equiv G_{Y}$.  The capacitor load is used for integrating the current proportional to the time derivative of the memresistor state variable $x$.
The corresponding LTSpice netlist code for memristor model denoted $K1$ is:
\begin{lstlisting}
1 .subckt K1 te be Y
2 .params ron = 100 roff = 16e3 k = 10e3 C1 = 1
3 C1 Y be IC = 0.3
4 R2 Y be 1G
5 Gy 0 Y value = (k V(te,be)(1/(ron(V(Y)) + roff(1 - V(Y))))(4V(Y)(1-V(Y)(1-V(Y))))}
6 G1 te be value = V(te,be)((1/(ron(V(Y)) + roff(1-V(Y)))))
7 End K1
\end{lstlisting}
Most memristors are based on transition metal oxides such as titantium dioxide, hafnium dioxide, and tantalum oxide.   Numerous titanium dioxide memristor nanostructure models have been proposed including Strukov-Williams, Joglekar-Biolek, Ascoli-Corinto, and Lehtonen-Laiho \cite{Mladenov:2021}   Memristive elements exhibit hystersis relationship between voltage and current.  Figure \ref{fig:structure}(a) illustrates the memristor nanostructure representing the metallic contacts (top electrode nad bottom electrode), the doped and un-doped layers of the memristor.  Figure \ref{fig:structure}(b) shows an equivalent circuit representing the resistances of the doped and the un-doped regions connected in series.

Memristors with diffusive dynamics are well suited to minmic the behavior of synapses as they can be scaled down to feature sizes below 10 nm, can retain memory states for years, switch with nanosecond timescales, and undergo spike-based learning in real time under biologically inspired learning rules as spike-time-dependent plasticity (STDP) \cite{Sune:2020}.    The characteristic I-V equations of a memristive element can be approximated by:
 \begin{align}
    i_{MR} &= G(w,v_{MR})v_{MR} \\
    dw/dt &= f_{MR}(w,v_{MR}) 
 \end{align}
 where $i_{MR}$ and $v_{MR}$ are the current voltage drop  at the terminal devices, respectively where
 \begin{equation}
    f_{MR} = \begin{cases}
                I_{0}\text{sign}(v_{MR})(e^{\lvert v_{MR} \rvert/v_{0}}-e^{v_{TH}/v_0}) \ \\ \ \ \ \ \ \ \ \ \ \ \ \ \ \ \ \ \ \ \ \ \ \ \ \ \text{if} \ \lvert v_{MR} \rvert > v_{TH} \\
                0 \ \text{otherwise}
             \end{cases}
 \end{equation}
 
 $G(w,v_{MR})$ is the conductance of the device that changes as function of the applied voltage (assuming a voltage or flux controlled device model), and $w$ is some physical parametric characteristic whose change is typically governed by a nonlinear function $f_{MR}$ of the applied voltage including a threshold barrier \cite{Camunas:2019}.  They are typically binary devices switching between high resistance state and low resistance state.

\begin{figure}[h]
	   \centering
	   %\begin{center} 
	   \includegraphics[width=0.8\linewidth]{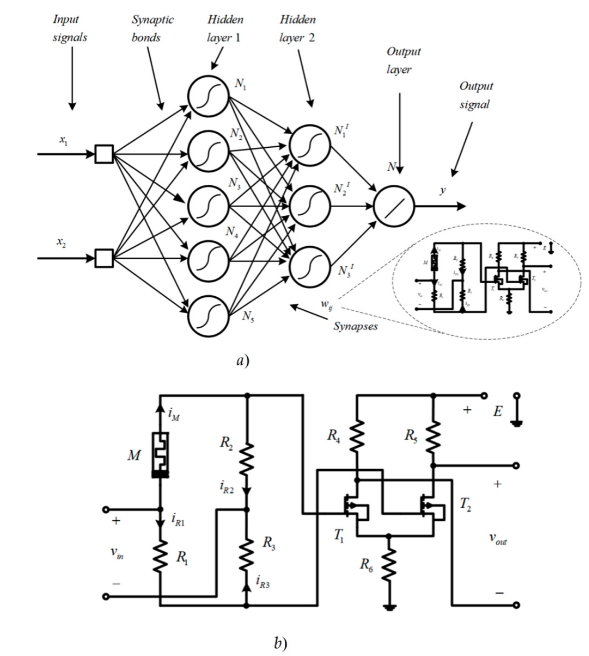}
	   %\small \caption{} 	
	   \caption{(a) Memristor neural network and (b) memristor-based synapse.  \cite{Mladenov:2021}}  
	   %\end{center}  
      \label{fig:nn} 
\end{figure}

\begin{figure}[h]
	   \centering
	   %\begin{center} 
	   \includegraphics[width=0.7\linewidth]{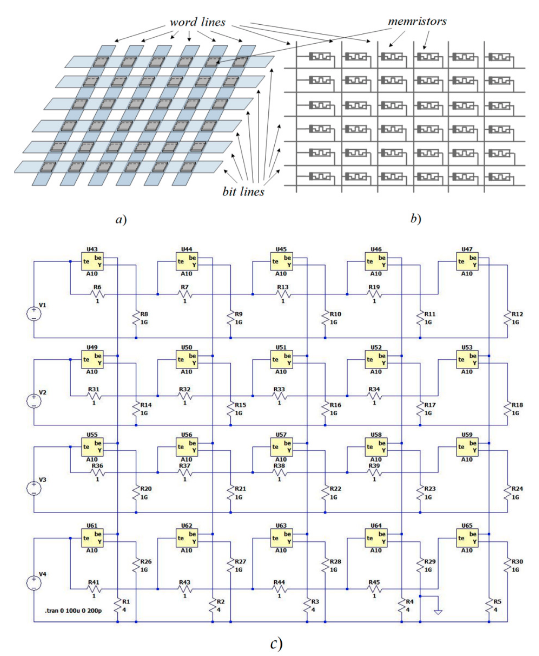}
	   %\small \caption{} 	
	   \caption{(a) Memristor memory crossbar; (b) equivalent circuit; (c) LTSpice schematic.  \cite{Mladenov:2021}}  
	   %\end{center}  
      \label{fig:bar} 
\end{figure}

 In hardware, a learned neural approximator can approximate the filter response with fewer dedicated fixed filter taps (fewer multipliers), and in memristor crossbar implementations the MAC is inherent in Ohm’s law (vectors multiplied by conductances and summed in analog) — potentially much lower area and power than digital MAC arrays. The tradeoffs: approximation error vs resource/power savings.  Figure \ref{fig:nn} shows the structure of a memresistor neural network and \ref{fig:nn}(b) shows a schematic of a memresistor-based synapse. 
  A schematic of a passive memristor memory crossbar is shown in Figure \ref{fig:bar}(a) and its equivalent circuit is shown in Figure \ref{fig:bar}(b).  The memristor memory crossbar schematic in LTSpice is shown in Figure \ref{fig:bar}(c).

\subsection{Time Registers and Operational Logic}

\begin{figure}[h]
	   \centering
	   %\begin{center} 
	   \includegraphics[width=0.5\linewidth]{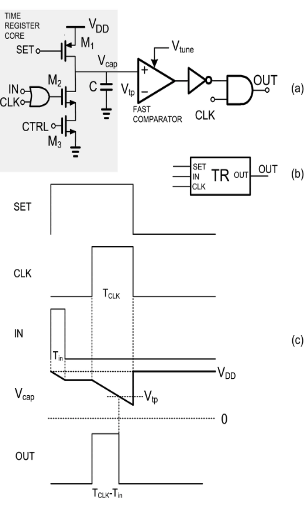}
	   %\small \caption{} 	
	   \caption{(a) Time register circuit; (b) symbol; and (c) timing diagram \cite{Felouris:2022}}  
	   %\end{center}  
      \label{fig:tr} 
\end{figure}

    The time register (TR) or shift register is a sequence of flip flops that stores bits and shifts them alone one position with each clock pulse, creating a time delay or performing serial-to-parallel/parallel-to-serial conversion, acting as temporary data storage synchronized by a clock. Figure \ref{fig:tr} shows the time register and its symbol.
    If SET = 0, transistor $M_{1}$, a PMOS, is turned on, and the capacitor voltage is set to the supply voltage $V_{D}$.   The capacitor discharges when the transistor $M_{2}$, an NMOS, is ON, which is controlled by an OR gate \cite{Felouris:2022}.
    
    Figure \ref{fig:tr} is a  time diagram of a time register that tracks the synchronization procedure.   The SET signal's time interval $T_{CLK}$ is a pulse with a fixed pulse width and a $25\%$ duty cycle. When  both IN and CLK are 0, then the capacitor voltage remains constant.   The larger the pulse width $T_{in}$ of the IN signal, the more discharging time due to $T_{in}$ appeared, considering that the discharging time due to CLK stays the same.   The output is a pulse with  a width equal to $T_{CLK} - T_{in}$, allowing the value of $T_{in}$ to be stored while the output pulse is synchronized with the CLK signal.

\begin{figure}[H]
	   \centering
	   %\begin{center} 
	   \includegraphics[width=1\linewidth]{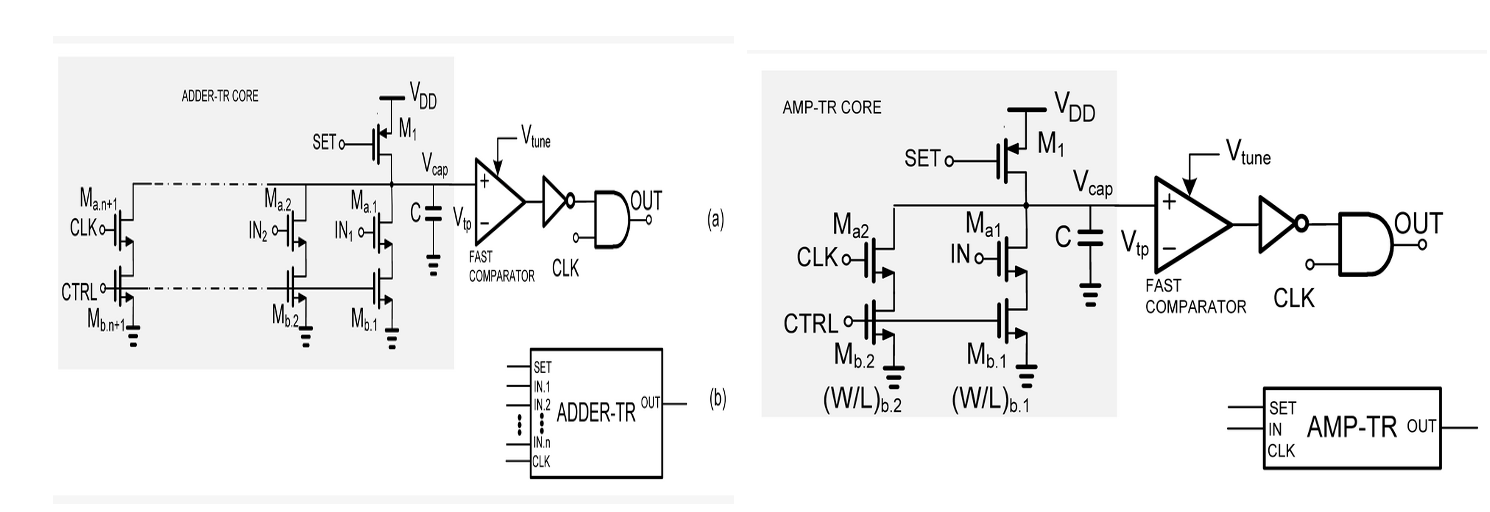}
	   %\small \caption{} 	
	   \caption{(left) Time amplifier on time register circuit and  symbol (ADDER TR); (right) time adder based on time register and symbol (AMP TR) \cite{Felouris:2022}}  
	   %\end{center}  
      \label{fig:op2} 
\end{figure}

    The circuit can store the time interval of an input pulse and amplify the pulse width by a gain factor. The new circuit is a time amplifier and is shown on the left in Figure \ref{fig:op2}.  Following Felouris et al. \cite{Felouris:2022}, in this configuration, the operation of the OR gate is performed by the two-transistor branches $M_{a1}–M_{b1}$ and $M_{a2}–M_{b2}$. Transistors $M_{a1}$ and $M_{a2}$ have the same aspect ratio acting as switches. The aspect ratios of $M_{b1}, M_{b2}$ are different, featuring a different discharging slope. Assuming that the channel widths of $M_{b2}$ and $M_{b1}$ are $W_{b,2}$ and $W_{b,1}$, respectively, while both transistors have the same channel length. Then, intuitively, using Figure \ref{fig:op}(c), the discharging slope between $T_{in}$ and $T_{CLK}$ will be different. The discharging slopein that corresponds to $T_{in}$ will be given by
 \begin{equation}
    slope_{in} = a \cdot slope_{clk}
 \end{equation}
where $a$ is the time gain and is given by
 \begin{equation}
    a = \frac{W_{b,2}}{W_{b,1}}
 \end{equation}
and $slope_{clk}$ is the discharging reference slope caused by $T_{CLK}$. Therefore, the output pulse width will be equal to
\begin{equation}
    T_{OUT} = T_{CLK} - \frac{W_{b,2}}{W_{b,1}}T_{in}
\end{equation}

    The right side of Figure \ref{fig:op2} shows a time adder circuit which is based on the time register.  A time adder simply adds the pulse widths $T_{in,1}, T_{in,2},\cdots, T_{in,n}$ of $n$ number input pulses. Transistors $M_{b,1}, M_{b,2},\cdots, M_{b,n}, M_{b,n+1}$ have the same aspect ratio \cite{Felouris:2022}. Therefore, the discharging slope caused by $T_{in,1}, T_{in,2},\cdots, T_{in,n}$ will be given by
 \begin{multline}
    slope_{in,1,in,2,\cdots,in,n} =  \\slope_{in,1} + slope_{in,2} + \cdots + slope_{in,n}
 \end{multline}
and the output pulse width will be 
\begin{equation}
    T_{OUT} = T_{CLK} - T_{in,1} - T_{in,2} - \cdots T_{in,n} 
\end{equation}
\subsection{Time-Domain Modules} 
\ \ DSP uses time lag/delay operations using a time mode $z^{-1}$ multiplier is to produce an output pulse with pulse $T_{out}$ width equal to $aT_{in}$ where $T_{in}$ is the input pulse and $a$ is the multiplication coefficient, and the output pulses will be synchronized with the sampling signal and delayed by one clock cycle.  A proposed $z^{-1}$ circuit is shown on the left in Figure \ref{fig:op} and the waveforms are shown on the right.  A combination of four TR circuits in series realizes a $z^{-1}$ circuit.   The SAMPLING CLOCK signal is assumed to be the SET1 signal of the TR1 circuit, and the input signal IN is equal to the input IN1 of TR1, while the final output signal OUT is the output of OUT4 of TR4 \cite{Felouris:2022}.

To be synchronized with the SET2 pulse, the OUT1 pulse is generated at the rising edge of CLK1. After that, the OUT1 was used as the TR’s input. OUT2 is synchronized with
CLK2, and $T_{OUT2} = T_{CLK} - T_{OUT1} = T_{CLK} - (T_{CLK} + T_{in}) = T_{in}$

\section{FPGA Design for DSP}
\label{FPGA}

IBM's TrueNorth Chip \cite{Akopyan:2015} and Intel's Loihi \cite{Davies:2018} are ASIC chips with neuromorphic architectures.  The TrueNorth core comprises five components: neuron block, core sram, router, scheduler, and token controller \cite{Valancius:2020}.    A single TrueNorth chip comprises 4,096 neurosynaptic cores.  Valancius et. al. \cite{Valancius:2020} prototypes
the TrueNorth architecture as a reference for a parameterized and configurable emulation platform for FPGA neuromorphic architecture on the Xilinx Zynq UltraScale+ MPSoC.  They validate the functionality of their emulation environment using the MNIST dataset and vector matrix multiplication case studies and demonstrate increased computational efficiency, reduced computing time, and a reduction in resource utilization while maintaining high accuracy.  A systematic review of digital neuromorphic architectures (NMAs) implemented on FPGAs is given in \cite{Szczerek:2025}.

\begin{figure}[h]
	   \centering
	   \includegraphics[width=0.75\linewidth]{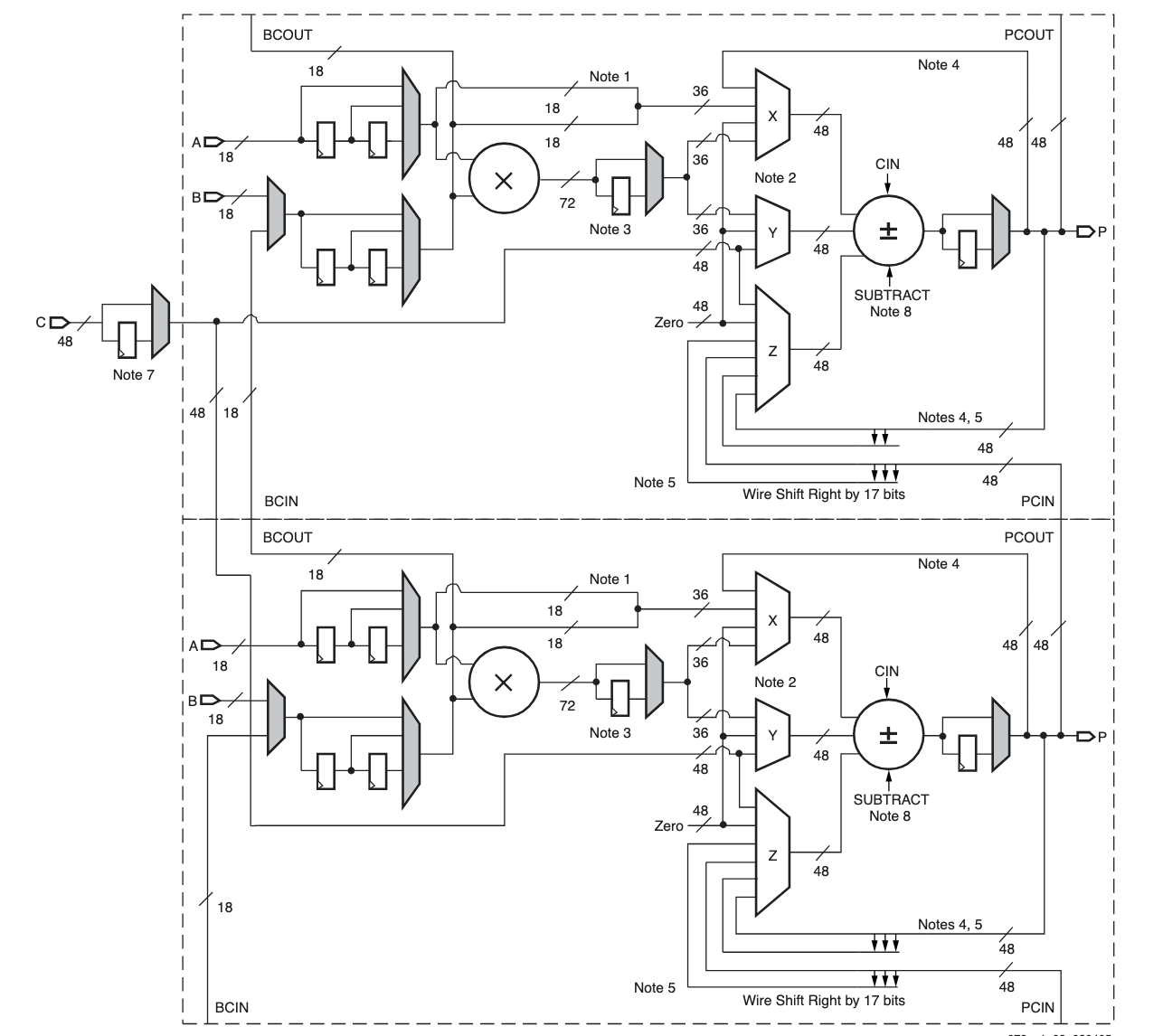}
	   \caption{DSP48 tile with two DSP48 slices. \cite{Xilinx:2005}}  
	   %\end{center}  
      \label{fig:xilinx} 
\end{figure}

FPGAs provide numerous multiply-accumulate (MAC) functions which are the fundamental building block of DSPs.  In contrast, a traditional DSP provides only one to four such MACs. The Virtex-4 family of Xilinx, built on an ASMBL architecture, contains a new element called the DSP48 slice that integrates high performance arithmetic and accumulation unit along with a multiplier.   DSP48 slices are cascadable into titles known as XtremeSDP tiles with two sets.  The DSP48 slice comprises four main sections: I/O registers, $18 \times 18$ signed multiplier, three input adder/subtractor blocks , and OPMODE multiplexers that support over 40 unique operations at 500 Mhz.  For DSP applications, the DSP48 modules would be used on the FPGA hardware.  Figure \ref{fig:xilinx}  shows a DSP48 tile consisting of two DSP48 slices.

The 18-bit A bus and A bbus are concatenated, wit the A bus being the most significant.
The X, Y, and Z multiplexers are 48-bit designs.   Selecting any of the 36-bit inputs provides a 48-bit sign-extended output.  The multiplier outputs two 36-bit partial products, sign extended to 48 bits.  The partial products feed the X and Y multipliers.  When OPMODE selects the multiplier, both X and Y multiplexers are utilized and the adder/subtracter combines the partial products into a valid multiplier result.    The multiply-accumulate path for P is through the Z multiplier.   The P feedback through the X multiplexer enables accumulation of P cascade when the multiplier is not used.  The grey-color multiplexers in \ref{fig:xilinx} are programmed at configuration time.  The shared C register supports multiply-add, wide addition, or rounding.  Enabling SUBTRACT implements 
Z - (X+Y+CIN) at the output of the adder/subtracter \cite{Xilinx:2005}.

\begin{figure}[h]
	   \centering
	   \includegraphics[width=0.75\linewidth]{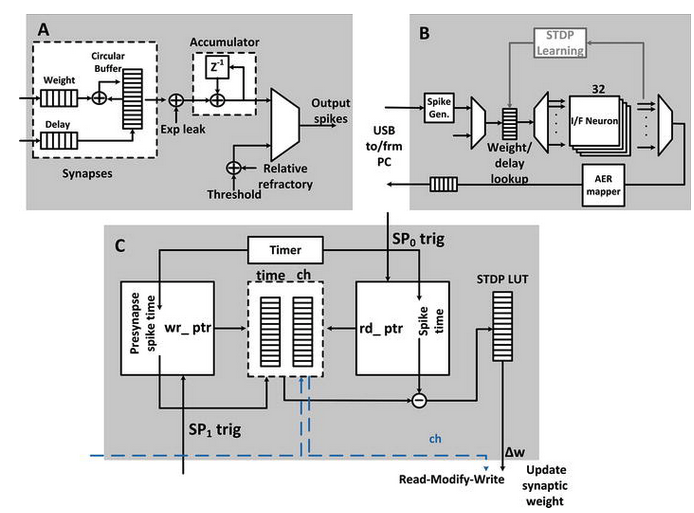}
	   \caption{FPGA-based neuron.  A: details the the individual neuron architecture; B: shows the overall network architecture including interfaces for spike generation; C: implementation of STDP 
      \cite{Cassidy:2007}}  
	   %\end{center}  
      \label{fig:na} 
\end{figure}

This DSP architecture can be modified with an FPGA-based neuron array design, such as proposed by Cassidy et al. \cite{Cassidy:2007} (illustrated in Figure \ref{fig:na}), that integrate spiking neuron networks using synapses, a delay mechanism, an accumulator, thresholds, and spike-time-dependent plasticity \cite{Mehrabi:2024}.  

Pearson et. al. \cite{Pearson:2007} propose an FPGA-based array processor architecture simulating large networks with more than a thousand neurons for real-time signal processing.   However, the scalability is limited by a bus-based communication protocol.  More advanced SNN architectures have been introduced for neuromorphic FPGAs including Optimized Deep Event-driven Spiking Neural Networks (ODESA) \cite{Mehrabi:2023}, Polychronous Spiking Neural Networks (PSNN) \cite{Izhikevich:2006, Wang:2013},  and  Embedded Multicore Building Blocks for Reconfigurable Architecture Computational Emulation (EMBRACE-FPGA) \cite{Morgan:2009, Cawley:2011, Mehonic:2024}

ODESA is a multi-layered architecture that employs  gradient-free, threshold-based learning algorithm that dynamically adjusts synaptic weights and neuronal thresholds based on local synaptic behavior and a Winner-Takes-All (WTA) mechanism.  Each neuron operates in an event-driven manner, processing inputs as discrete binary spikes, significantly reducing power consumption and processing.  In PSNNs, spikes from different neurons travel down axons with precise delays, arriving simultaneously at a target neuron and causing it to fire, even though the source neurons fire asynchronously.
Instead of employing traditional analog neurons, EMBRACE models these neurons using soft-processors on the FPGA, facilitating a novel approach to neural computation that leverages the reconfigurability of digital systems.

\section{Results}
\label{results}
We simulate a sinusoidal signal plus Gaussian noise using a signal amplitude $A = 0.6$, a frequency $f = 50$ Hz, a noise amplitude $A_{noise} = 0.05$, and then quantized to fixed point Q15.
\begin{equation}
x[n] = A\text{sin}(2 \pi f n) + 2A_{noise}(\text{rnd}\%1000)/1000 - 1)  \nonumber
\end{equation}
where $\text{rnd} \% 1000$ is a Gaussian random number generator mod 1000 and $n$ is the time step from $[0,2000)$.  The signal is simulated in both a training phase and then compared in a testing phase each with 2000 time steps.  We implement a classic FIR, classical IIR, neuromorphic FIR, and neuromorphic IIR models.  The neuromorphic FIR model is small feedforward NN (single hidden layer) that approximates the linear FIR convolution.  
We use a classical FIR: direct-form, parameterized taps, fixed-point (WIDTH bits, FRAC fractional bits) and a classical IIR direct Form II transposed (numerical stability), parameterized order and fixed-point.   The neuromorphic IIR uses a small recurrent neural network (Elman-type) that approximates the IIR filtering behavior and implemented with a recurrent vector multiply and activation (tanh approximated by LUT).  

\begin{figure}[h]
	   \centering
	   %\begin{center} 
	   \includegraphics[width=1\linewidth]{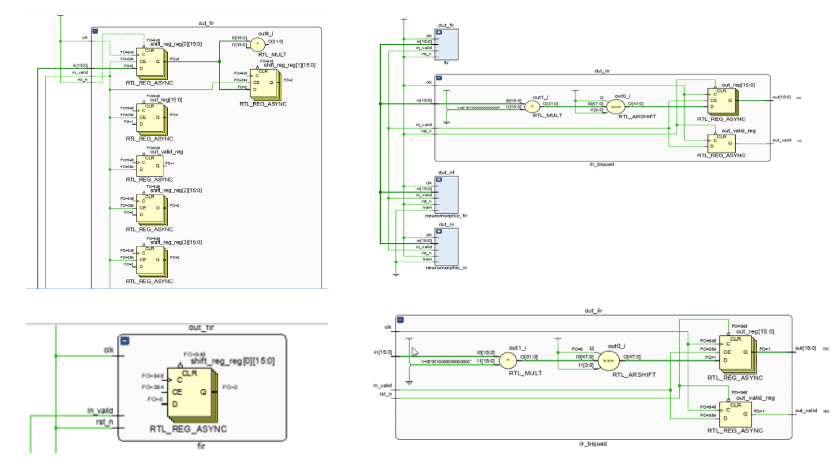}
	   %\small \caption{} 	
	   \caption{Schematics. (left) FIR and (right) Biquad IIR}  
	   %\end{center}  
      \label{fig:nm} 
\end{figure}

\begin{figure}[h]
	   \centering
	   %\begin{center} 
	   \includegraphics[width=1\linewidth]{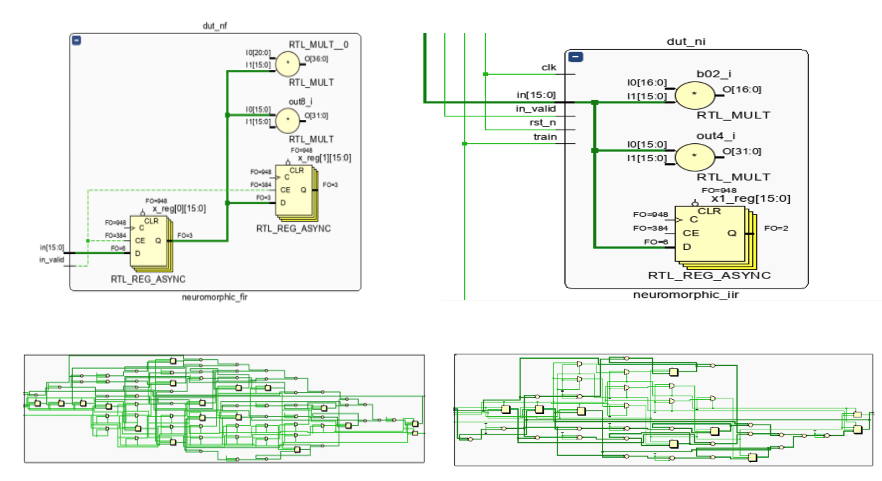}
	   %\small \caption{} 	
	   \caption{Schematics. (left) neuromorphic FIR and (right) neuromorphic Biquad IIR}  
	   %\end{center}  
      \label{fig:nm2} 
\end{figure}

\begin{figure}[H]
	   \centering
	   %\begin{center} 
	   \includegraphics[width=1\linewidth]{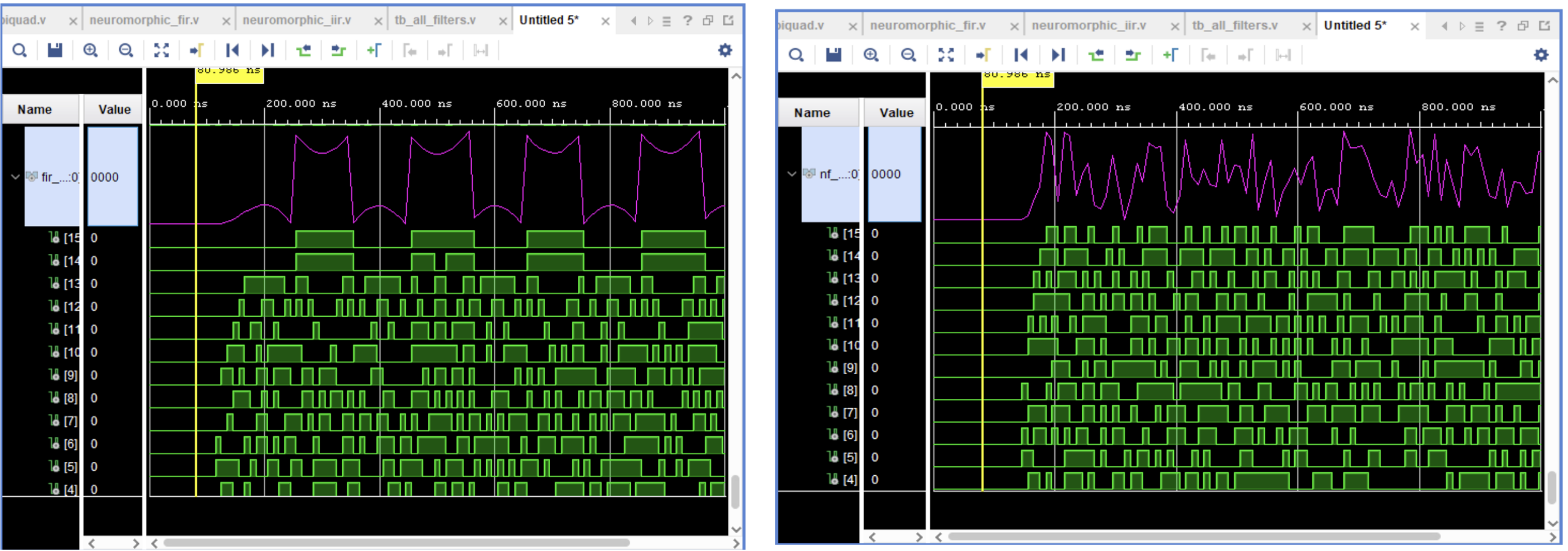}
	   %\small \caption{} 	
	   \caption{(left) Classical FIR And (right) Neuromorphic FIR}  
	   %\end{center}  
      \label{fig:nfir} 
\end{figure}

%We use fixed-point format: signed Q(1) (DATA WIDTH-1) (one sign bit + fractional bits). Verilog uses synchronous multiply-accumulate (MAC) using integer arithmetic; coefficients and data are fixed-point integers. Outputs are scaled appropriately.  
Neuromorphic versions emulate a memristor weight array in logic (weights stored as registers) and include an LMS learning rule inside the module (for online adaptation). The learning rule is only invoked when \code{train}=1 and a desired signal is provided by the testbench.  We IIR uses Direct Form I biquad (2nd order) with fixed-point coefficients; The neuromorphic IIR uses an adaptive recurrent weight structure with LMS updates for feedback/feedforward taps.  Testbenches provide the same stimulus (sinusoid + noise, impulse, step) to all variants so output comparison is fair.  LTSpice netlists implement crossbar sums using memristors and an op amp current summing node. The memristor behavior is a simple physics-inspired behavioral model (suitable for performance testing). %Replace with a published memristor model when needed.
Schematics generated in Vivado for the FIR and Biquad IIR models are shown in Figure \ref{fig:nm}.
Schematics for the neuromorphic FIR and the neuromorphic IIR Biquad are shown in Figure \ref{fig:nm2}.
Inputs are tapped samples. In practice, weights are trained offline and loaded into hardware registers (or an on-chip memristor array in LTSpice). It can use fewer multipliers if one quantizes \& re-use weights or exploit sparsity. 
The following table compares the MSE of the models relative to the classical FIR golden reference.  Thus, the MSE for the classical FIR is 0.
\begin{table}[h]
\centering
\caption{MSE relative to classical FIR golden reference}
\begin{tabular}{c c c c c}
\hline 
    Model & MSE & Cells & Cell Pins & NetLists \\
\hline
    Classical FIR  & 0.0  & 173  & 36 & 921 \\
    Neuromorphic FIR & 0.5209 & 335 & 53 & 1544\\
    Classical IIR & 0.1374 & 192 & 36 & 1007\\
    Neuromorphic IIR & 0.1533 & 163 & 53 & 760\\
\hline
\end{tabular}
\end{table}
    Figure \ref{fig:nfir} shows the simulated output of the classical FIR (left) and the neuromorphic FIR (right).  As shown, the neuromorphic FIR is much more variable than the classical FIR. Neuromorphic Finite Impulse Response (FIR) signals are much more variable than classical FIR signals primarily due to the inherent noise and component variations in the underlying neuromorphic hardware, which often uses analog or mixed-signal circuits to mimic biological processes. Classical digital signal processing (DSP), by contrast, operates on precise numerical values in a highly controlled, deterministic environment.
\begin{figure}[H]
	   \centering
	   %\begin{center} 
	   \includegraphics[width=1\linewidth]{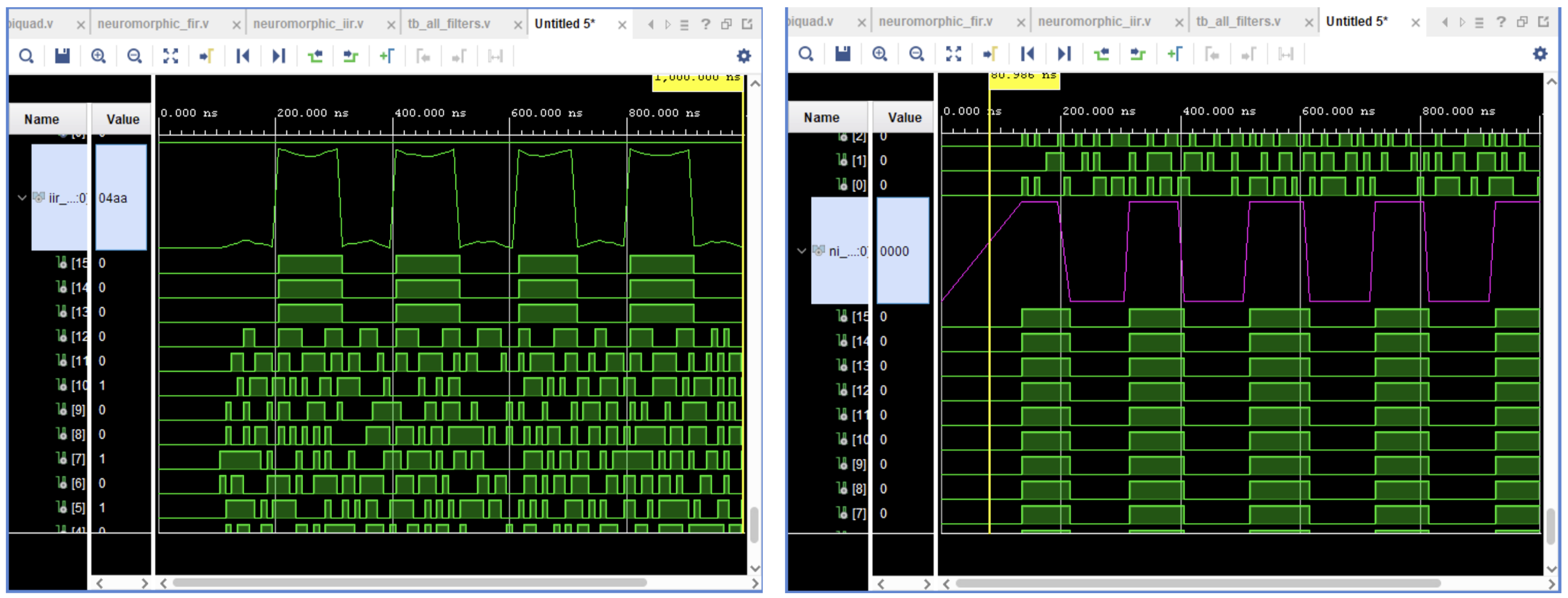}
	   %\small \caption{} 	
	   \caption{(left) Classical IRR and (right) Neuromorphic IIR}  
	   %\end{center}  
      \label{fig:three} 
\end{figure}

\section{Conclusion}
\label{conclusion}
    Using Vivado and Verilog HDL for FPGA design, FIR and IIR filters using neuromorphic computing were simulated and compared to standard FIR and IIR von Neumann-based filters.  Spiking neural networks were used to generate event-based signals.  In actual hardware implementation, memristors in DSP48 modules would be used.   IIR neuromorphic computing and standard IIR filter have comparable MSEs relative to the FIR golden.  The FIR neuromorphic model relative to the standard FIR golden reference has much higher MSE.  Though neuromorphic computing may not provide as precision results as standard DSP that uses sequential logic, it does have significant power consumption, efficiency, and cost benefits.  Furthermore, the distributed nature of neuromorphic networks makes them inherently more robust to faults or errors.   Due to its parallel nature and adaptive learning while integrating processing and memory within the same location, similar to how neurons and synapses work in the brain, neuromorphic computing eliminates the von Neumann bottleneck.  Future work will involve research into improving precision for real-time DSP applications and actual FPGA implementation using Xilinx.
\printbibliography

\end{document}